\documentclass[lettersize,journal]{IEEEtran}
\usepackage{amsmath,amsfonts}
\usepackage{dsfont}
\usepackage{algorithm}
\usepackage{algorithmic}
\usepackage{setspace}
\usepackage{array}
\usepackage{textcomp}
\usepackage{stfloats}
\usepackage{url}
\usepackage{verbatim}
\usepackage{graphicx}
\usepackage{subfigure}
\usepackage{booktabs}
\usepackage{multirow}
\usepackage{natbib}
\def\BibTeX{{\rm B\kern-.05em{\sc i\kern-.025em b}\kern-.08em
    T\kern-.1667em\lower.7ex\hbox{E}\kern-.125emX}}
\usepackage{balance}
\begin{document}
\title{Time Tracker: Mixture-of-Experts-Enhanced Foundation Time Series Forecasting Model with Decoupled Training Pipelines}
\author{Aobo Liang, Xiaohou Shi, Ke Liand Yan Sun
\thanks{A. Liang and Y. Sun are with the School of Computer Science (National Pilot Software Engineering School), Beijing University of Posts and Telecommunications, Beijing 100876, China. E-mail: liangaobo@bupt.edu.cn; sunyan@bupt.edu.cn.}
\thanks{S. Xiao and K. Li are with the China Telecom Research Institute, Beijing 102209, China. E-mail: shixh6@chinatelecom.cn; lik24@chinatelecom.cn.}

}

\markboth{Journal of \LaTeX\ Class Files,~Vol.~18, No.~9, September~2020}%
{How to Use the IEEEtran \LaTeX \ Templates}

\maketitle

\begin{abstract}
In the past few years, time series foundation models have achieved superior predicting accuracy. However, real-world time series often exhibit significant diversity in their temporal patterns across different time spans and domains, making it challenging for a single model architecture to fit all complex scenarios. In addition, time series data may have multiple variables exhibiting complex correlations between each other. Recent mainstream works have focused on modeling times series in a channel-independent manner in both pretraining and finetuning stages, overlooking the valuable inter-series dependencies. To this end, we propose \textbf{Time Tracker} for better predictions on multivariate time series data. Firstly, we leverage sparse mixture of experts (MoE) within Transformers to handle the modeling of diverse time series patterns, thereby alleviating the learning difficulties of a single model while improving its generalization. Besides, we propose Any-variate Attention, enabling a unified model structure to seamlessly handle both univariate and multivariate time series, thereby supporting channel-independent modeling during pretraining and channel-mixed modeling for finetuning.
Furthermore, we design a graph learning module that constructs relations among sequences from frequency-domain features, providing more precise guidance to capture inter-series dependencies in channel-mixed modeling. Based on these advancements, Time Tracker achieves state-of-the-art performance in predicting accuracy, model generalization and adaptability.
\end{abstract}


\section{Introduction}
\IEEEPARstart{R}{ecent} advances in time series forecasting have demon-strated the effectiveness of foundation models, with Transformers emerging as the pivotal architecture. With the auxiliary information of endogenous, exogenous variables and cross-domain features from historical contexts, time series foundation models are designed to accommodate a wider spectrum of prediction scenarios. The concept of unified forecasting is gradually reshaping the conventional practice of task-specific training strategy.

\begin{figure}[t]
    \centering
    \includegraphics[width=1.0\linewidth]{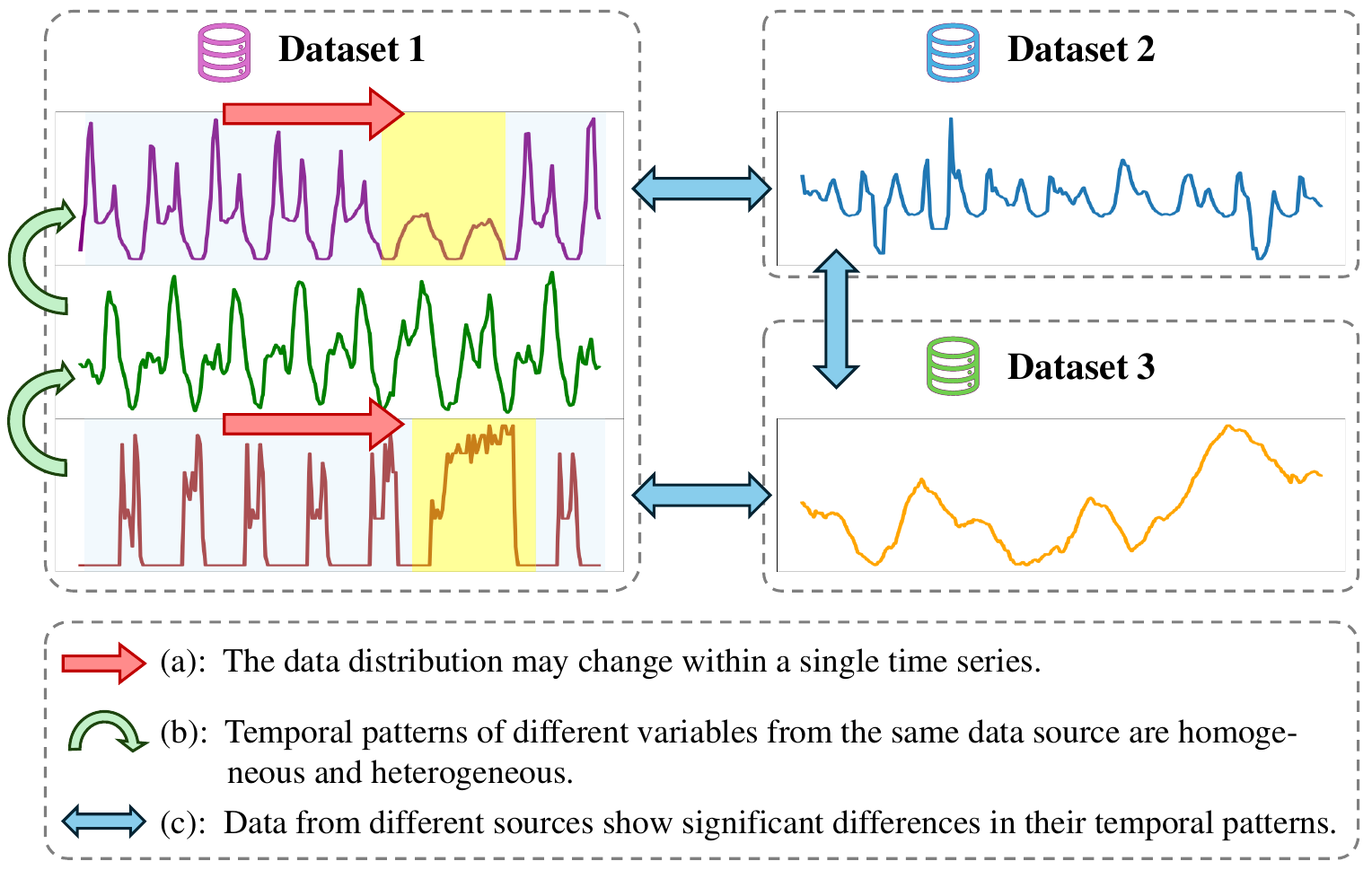}
    \caption{Homogeneous or heterogeneous time series may exhibit differences in data distribution across (a) different time intervals within a single sequence, (b) varying sequences from the same data source and (c) time series from different data sources.}
    \label{fig:fig1}
\end{figure}

However, existing models face performance bottlenecks in multivariate forecasting. Firstly, as shown in Fig. \ref{fig:fig1}, real-world time series are inherently heterogeneous: (a) within a single sequence, and across different variables from (b) the same or (c) different scenarios. The complex temporal patterns and shifting data distributions pose significant challenges for a single model to extract high-quality features. Meanwhile, recent research has mainly focused on building encoder-only architectures to model temporal dependencies across tokens. Although these models often achieve higher accuracy on specific datasets, they tend to suffer from limited generalization. In contrast, generative decoder-only architectures offer better scalability and are becoming an emerging focus of research. Moreover, multivariate time series usually exhibit complex inter-series dependencies, including immediate or delayed correlations in trends and seasonality. Most previous works adopt channel-independent modeling for both pretraining and finetuning stages, where sequences from different variables are processed independently. Considering the significant variation in the number of variables across different real-world applications, it is impractical to rely on a single model structure for all possible multivariate forecasting tasks. 
Nevertheless, it is undeniable that the channel-independent strategy overlooks valuable inter-variable dependencies, which may become a bottleneck limiting the model’s adaptability on specific datasets. Some recent works, such as Moirai\citep{moirai} and Timer-XL\citep{liu2024timerxl}, have established a unified training paradigm to support multivariate pretraining and finetuning. However, these models try to establish relations among tokens from all input variables without any filtering mechanisms. With the assumption that all input sequences are interrelated, the model may face increased learning cost and introduce noise to tokens from weakly correlated or unrelated variables.

To address the aforementioned issues, we propose \textbf{Time Tracker}, a novel generative time series forecasting model architecture. To tackle the problem of highly heterogeneous distributions of time series, we are inspired by sparse mixture of experts (MOE) to assign time series with diverse distributions to refined experts. 
Unlike previous token-level MoE approaches\cite{shi2024timemoe} where individual tokens are dispatched to different experts, we assigns all tokens originating from a single time series to the same expert. This sequence-level routing mechanism enables each expert to learn the underlying evolution dynamics of each series, thereby enhancing its ability to model coherent long-term patterns. In addition, we propose Any-variate Attention (AVA) to enable a unified model structure to seamlessly handle both univariate and multivariate time series causally. Typically, a well pretrained time series foundation model should be equipped with generalizable parameters by training it on data with diverse distributions, achieving robust performance across varied tasks and domains. Considering that the variable numbers and the relations among them may vary significantly across application scenarios, it is inherently challenging to generalize the inter-series dependencies of different domains. Therefore, we adopt the channel-independent strategy during pretraining to ensure the model's generalization capacity. 

To further adapt the pretrained model to unseen datasets, we perform finetuning stage which in contrast transfers AVA to capturing data-specific inter-series dependencies. We first design a frequency-based graph learning layer to compute the relationship probability between the variables, then use the reparameterization technique to generate the adjacency matrix. We inject such relational information into AVA through the Kronecker product of the adjacency matrix and the causal attention time mask matrix. Consequently, the model can precisely control the interactions of tokens between different variables in a generative manner. In summary, our main contributions are as follows:
\begin{enumerate}
    \item We propose to assign sequence tokens originating from the same series to the same expert networks to reduce the learning difficulty of specific data distributions and improve prediction performance.
    \item We propose to combine the context-aware graph neural network with causal attention to better adapt to the downstream tasks of multivariate metric data. While the model works in a generative manner within different variables, it captures multivariable dependencies more accurately.
    \item We pretrain our model on the Unified Time Series Dataset (UTSD) and conduct experiments on multivariate forecasting, zero-shot learning and few-shot learning. Timer Tracker achieves SOTA performance compared to other benchmark models.
\end{enumerate}

\section{Related Work}
\subsection{Large Time Series Models}
Recent advances in pre-training on large-scale sequence data have significantly benefited modality understanding in natural language processing\citep{grattafiori2024llama} and computer vision\citep{liu2021swin,kirillov2023segment}. The similar trends has been extended to time series modeling. However, most methods\citep{wu2021autoformer,zeng2023transformers,patchtst} are limited in model scale or in-domain applicability, which results in weak generalization ability. When encountering temporal patterns or data distributions previously unseen, these models often require re-training or extensive fine-tuning, significantly increasing deployment cost and reducing practical scalability. To address this limitation, LTSM\citep{liu2024timer} are being explored through large-scale pre-training to enhance zero-shot generalization across diverse scenarios. A key challenge lies in managing the inherent heterogeneity of time series data, including variations in domain, frequency and semantics. Some approaches introduce tokenization reprogramming framework\citep{chen2024model} and attempt to leverage the generalization capability of existing LLMs to adapt to the data distribution of time series. Time-LLM\citep{jintime} generates prompt embeddings by manually crafting dataset-specific prompts and proposes a reprogramming method to map time series into text prototypes, enabling direct prediction using existing LLMs. Lag-Llama\citep{rasul2023lag} maps time series into text prototypes by combining sequence samples at specific time lags with timestamps at multiple time intervals. Chronos\citep{ansarichronos} tokenizes time series into discrete bins through simple scaling and quantization of real values, achieving time series forecasting through minimizing the loss of a classification task. However, these methods are either reliant on the quality of training datasets or require manual specification of prompts or data sampling rules, which results in higher training costs. Alternatively, recent studies focus on model architectures designed specifically for time series forecasting. MOMENT\citep{goswami2024moment} follows the paradigm of pre-trained NLP models such as BERT\citep{devlin2019bert} by employing a reconstruction-based objective to train a Transformer Encoder as a feature extractor, with task-specific output heads designed for different downstream tasks. Although MOMENT is a general-purpose model tailored for time series analysis, it still requires fine-tuning for specific downstream tasks and cannot be directly applied to forecasting tasks. TimesFM\citep{das2024decoder} employs randomly sized masks to enable training with variable-length inputs to fixed-length outputs. Timer\citep{liu2024timer} performs next-token prediction and uses the causal attention mechanism to model time series in an autoregressive manner. Nevertheless, these models process time series in a channel-independent way and overlook the inter-series dependencies among different variables in both pre-training and fine-tuning stages. To address this issue, Moirai\citep{moirai} proposes a unified training strategy that allows for multivariate time series predictions. However, Moirai implicitly assumes that the input variables from the same data source are mutually correlated, which may introduce noise to each token due to capturing the dependencies of heterogeneous sequences.
\subsection{Mixture of Experts}
Mixture of Experts (MoE) has emerged as a solution to scale model capacity efficiently without proportionally increasing computational costs. Through dynamically routing inputs to specialized subnetworks, MOE enables the construction of large-scale models with sparse and resource-efficient computation. Recent works like Time-MOE\citep{shi2024timemoe} and Moirai-MOE\citep{liu2024moirai} assign different tokens to distinct experts to enhance the scalability and convergency of the original models. However, the in-depth motivation of the token-wise routing approach is ambiguous and ignore the statistical information of the original time series. In addition, existing works follow the auxiliary-loss-based load-balance strategy in Switch transformers\citep{fedus2022switch} which may lead to suboptimal performance due to imbalanced loss weight assignment and potential difficulties in achieving equitable expert utilization\citep{dai2024deepseekmoe}.
\subsection{Graph Neural Networks}
GNN has been widely used for spatial temporal modeling in multivariate forecasting. Previous works such as STGCN and Graph Wavenet use GNN to model sensors of different roads as nodes in the graph structure. Although the predicting performance is improved, the model needs predefined graph structure, which is usually not achievable in more complex scenarios. MTGNN introduces an adaptive graph learning layer to automatically generate an optimal adjacency matrix tailored to specific data. However, when faced with new scenarios involving different nodes, the model requires retraining, which limits its scalability. To address this issue, STEP proposes an instance-wise graph learning module to adapt to downstream tasks with any possible variables. STEP combines Bernoulli sampling with gumbel softmax so that the model can generate adjacency matrix based on the similarity between embeddings of different time series.

\begin{figure*}
    \centering
    \includegraphics[width=0.9\linewidth]{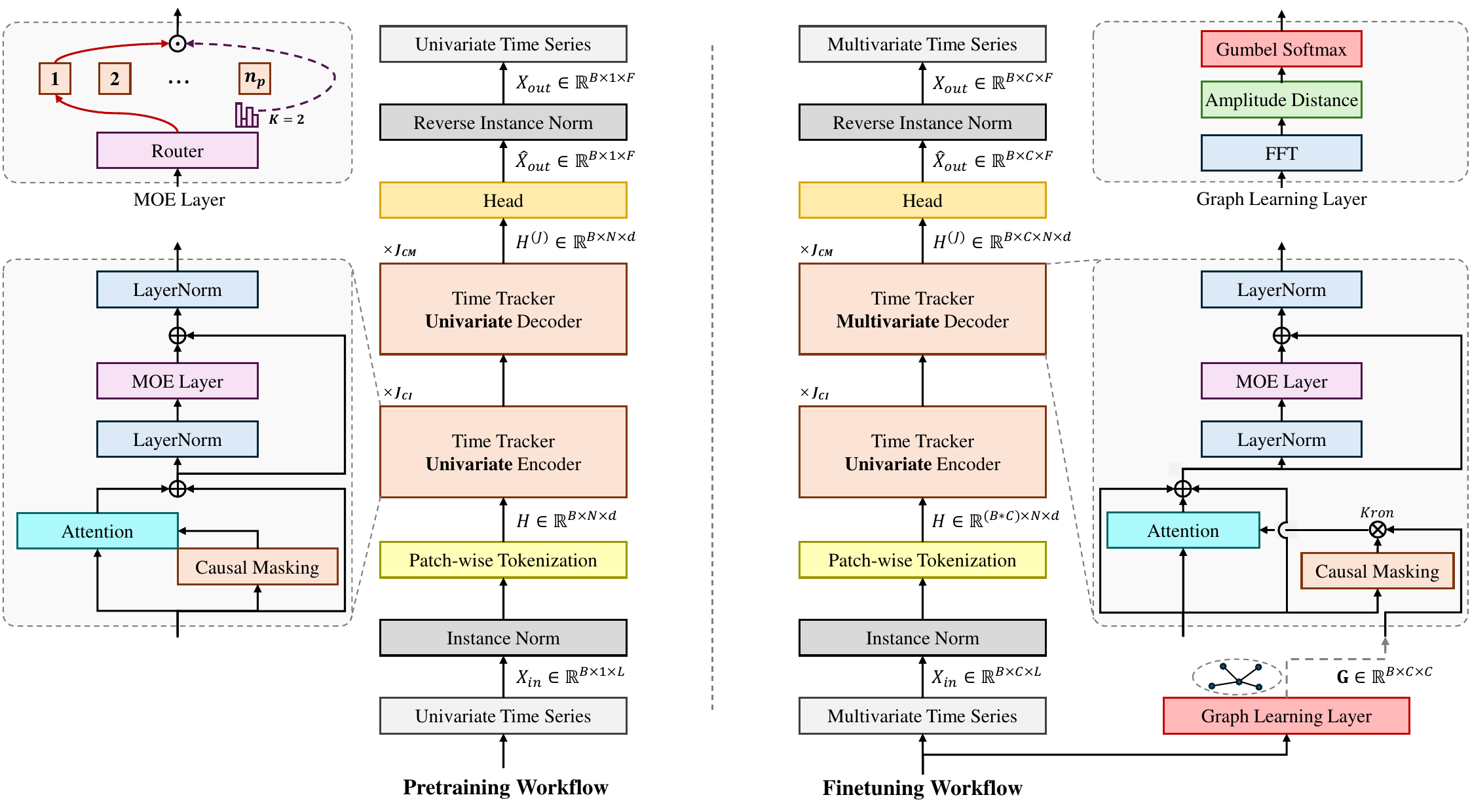}
    \caption{The model architecture of Time Tracker. The workflow of pretraining stage and finetune stage is on the left and right side respectively.}
    \label{fig:arc}
\end{figure*}

\section{Methodology}
In this section, we propose Time Tracker, a fundamental time series model built for multivariate forecasting. As shown in Fig \ref{fig:arc}, Time Tracker is based on a decoder-only Transformer architecture. We pretrain the model in the channel-independent manner where each univariate time series is fed into the model separately. In this setup, all model parameters are shared across different time series to capture diverse temporal evaluation patterns, therefore enhancing its generalization ability to unseen or heterogeneous time series data. To achieve more accurate prediction, we argue to capture inter-series dependencies while applying the model on specific datasets. In the fine-tuning stage, we design an adaptive graph learning layer to generate instance-wise adjacency matrices. We further combine the causal mask with the adjacency matrix to capture inter-series dependencies while keeping the autoregressive predicting scheme.

\subsection{Patch-wise Next Token Prediction}
Considering the input multivariate time series data $X_{in}=\{x_{1,t},x_{2,t},...,x_{C,t}\}_{t=1}^{L}$, our objective is to predict the future values $X_{out}=\{x_{1,t},x_{2,t},...,x_{C,t}\}_{t=L+1}^{L+F}$, where $C$ denotes the variable dimension, $L$ is the length of historical look-back window and $F$ represents the forecasting horizons. We first segment each series into $X_p=\{p_{1,\tau},p_{2,\tau},...,p_{C,\tau}\}_{\tau=1}^{N}$ where each patch $p_{i,j}=x_{i,(j-1)P+1:jP}$ corresponds to $P$ consecutive time points from the input series. The patches are extracted using a sliding window with stride $S$, producing $N=\lceil \frac{L-P}{S}+1 \rceil$ patch-wise tokens. We set $F=P$ so that the model learns to predict the next patch of length $P$ given the previous $N$ patches. We pass each token in $X_p$ through a linear projector to form multivariate patch-wise tokens $H=\{h_{1,\tau},h_{2,\tau},...,h_{C,\tau}\}_{\tau=1}^{N}$ for further processing, where $h_{i,j}\in\mathbb{R}^{d}$ and $d$ is the token dimension.

\subsection{Decoupled Training Pipelines}
\subsubsection{Pretraining Workflow}
Pretrained models are designed to capture generic and transferable patterns from large-scale data, enabling better generalization across different tasks and domains\citep{devlin2019bert}. In the context of time series forecasting, modeling each series independently allows the model to focus on the intrinsic temporal dynamics of individual variables, which facilitates the learning of varying temporal patterns and can improve generalization across diverse series \citep{patchtst}. Besides, multivariate time series from different scenarios often exhibit significant discrepancies in both dimensionality and the underlying inter-series relationships. Modeling such heterogeneous dependencies in a unified pretraining framework poses challenges in terms of scalability and representation consistency. Therefore, we adopt a channel-independent data loading strategy in the pretraining stage. Given a training dataset $\mathcal{D}$ with $num_\mathcal{D}$ multivariate subsets, we uniformly decompose these subsets into univariate time series and extract training samples accordingly. Specifically, let $\mathcal{D}_i\in\mathbb{R}^{C_i\times T_i}$ denote the training segment of the $i$-th subset with $C_i$ variables of length $T_i$. We generate training instances of the shape $1\times (L+F)$. Accordingly, the total number of training samples generated from $\mathcal{D}$ is $\sum_i^{num_{\mathcal{D}}}C_i\times (T_i-L-F)$.

\subsubsection{Finetuning Workflow}
Fine-tuning refers to re-training a pretrained model using data from specific tasks for domains, in order to make the model better suited for the target scenario. Previous works often employ consistent data loading strategies across pretraining and fine-tuning, preserving the univariate forecasting paradigm for the model. Although such a design has been widely adopted for its simplicity and scalability, it ignores the complex relationships among multivariate series. Incorporating inter-series relationships allows the model to leverage shared dynamics across variables. Such shared structure provides additional constraints that may reduce uncertainty when forecasting an individual series. Inspired from iTransformer\citep{itransformer}, we argue that capturing inter-series dependencies of multivariate data in specific datasets could provide better adaptability. This motivates us to introduce a graph learning module in the fine-tuning stage, estimating pairwise similarities among variables and represent the learned relationships as an adjacency matrix $\mathbf{G}\in\mathbb{R}^{C\times C}$. When modeling inter-series dependencies, $\mathbf{G}$ serves as a guiding for establishing the attention mask, which will be described in detail in Section \ref{sec::attention}.

Additionally, prior works such as LoRA\citep{hu2022lora} have shown that fine-tuning a restricted set of parameters can also achieve competitive performance. Motivated by this observation, we freeze the first $J_{CI}$ layers and update only the last $J_{CM}$ layers during finetuning, where $J_{CI}+J_{CM}=J$ and $J$ refers to model layers. For a batched input data $X_{in}\in\mathbb{R}^{B\times C\times L}$, we first reshape it into the shape of $\mathbb{R}^{(B*C)\times 1\times L}$ to maintain univariate processing in the first $J_{CI}$ layers, producing $H^{(J_{CI})}\in\mathbb{R}^{(B*C)\times 1\times N\times D}$. Consequently, we concatenate the tokens to $\tilde{H}^{(\ell_{CM})}\in\mathbb{R}^{B\times (C*N)\times D}$ and feed it into the last $J_{CM}$ layers. Through the adjacency matrix and any-variate causal attention layers in Sec\ref{sec::graph} and Sec\ref{sec::attention}, we capture inter-series dependencies in a causal manner to achieve better adaptation.

\subsection{Frequency-based Adaptive Graph Learning Layer}\label{sec::graph}
To capture the underlying relationships among different time series, we explore the similarity between different series in the frequency domain. Unlike direct modeling in the time domain, this approach allows us to extract invariant spectral features that are more robust to localized variations and more intuitive in terms of representing periodic patterns. Therefore, it facilitates the learning of more stable and generalizable inter-series relationships. For any given multivariate samples $X_{in}\in\mathbb{R}^{C\times L}$, we apply the real fast Fourier transform (RFFT) to obtain $X_{in}^{\mathcal{F}}\in\mathbb{R}^{C\times\frac{L}{2}}$. Subsequently, we compute the differences in amplitude values across various frequencies to derive a probability matrix $\mathbf{P}$ that quantifies the inter-series correlations, as shown in eq \ref{eq::freq}:
\begin{equation}\label{eq::freq}
\begin{aligned}
X_{in}^{\mathcal{F}}&=\mathrm{RFFT}(X_{in})=\{x_{1,t}^{\mathcal{F}},x_{2,t}^{\mathcal{F}},...,x_{C,t}^{\mathcal{F}}\}_{t=1}^{\frac{L}{2}}\\
\mathbf{D}_{ij}&=\left| x_{i,t}^{\mathcal{F}}-x_{j,t}^{\mathcal{F}} \right|\\
\tilde{\mathbf{Z}}_{i,j}&=\left(\sum\nolimits_{t=1}^{\frac{L}{2}}\alpha_{t}\log\left(1+\mathbf{D}_{ij}\right)\right)^{-1}\\
\mathbf{Z}_{ij}&=\begin{cases}
\mathrm{Sigmoid}\left(\frac{\mathbf{Z}_{ij}-\mu_{\Omega}}{\sigma_{\Omega}}\right)&\text{if }i\ne j\\
1&\text{if }i=j
\end{cases}
\end{aligned}
\end{equation}
where $\mathbf{D}$ is the distance matrix, $\mathbf{Z}$ represents similarity score matrix, $\alpha\in(0,1)$ is a learnable weighting factor that controls the relative importance of different frequency components, $\mu_{\Omega}$ and $\sigma_{\Omega}$ denote the mean and standard deviation computed over the off-diagonal entries of $\tilde{Z}$. The logarithmic transformation alleviates scale imbalance in frequency-domain distances induced by data distribution heterogeneity, thereby preventing large-magnitude components from dominating the similarity estimation. Subsequently, Z-score normalization and sigmoid are applied to the off-diagonal entries to generate similarity score matrix $Z$. This operation standardizes relative similarities and improves the stability of the inferred inter-series dependency structure.

Given the similarity score matrix $\mathbf{Z}$, we construct a variable correlation adjacency matrix $\mathbf{G}$ via Bernoulli resampling. A higher $\mathbf{Z}_{ij}$ indicates stronger probabilities of connection between variable $i$ and $j$. Since $\mathbf{Z}$ depends on learnable parameters $\alpha$, naive sampling would block gradient flow and prevent end-to-end optimization. Therefore, we adopt the Gumbel-Softmax reparameterization trick\citep{jang2017categorical}, which allows differentiable sampling during the Bernoulli resampling process. As Bernoulli sampling requires a binary-valued probability parameter, we first transform each entry of $\mathbf{Z}$ into two-class logits via the logit function, i.e., $\big[\log(\frac{\mathbf{Z}}{1-\mathbf{Z}}), \log(\frac{1-\mathbf{Z}}{\mathbf{Z}})\big]$. we then introduce a random value $\epsilon\in\mathbb{R}^2$ sampled from $\mathrm{Gumbel}(0,1)$ distribution and compute $\mathbf{G}$ through eq \ref{eq::gumbel}, where $\tau$ is the temperature parameter which control the sharpness of distribution to achieve approximate Bernoulli sampling.

\begin{equation}\label{eq::gumbel}
\begin{aligned}
\mathbf{P}_{ij}&=\big[\log(\frac{\mathbf{Z}_{ij}}{1-\mathbf{Z}_{ij}}), \log(\frac{1-\mathbf{Z}_{ij}}{\mathbf{Z}_{ij}})\big]\\
\mathbf{G}_{ij}&=\mathrm{Softmax}((\mathbf{P}_{ij}+\epsilon)/\tau)
\end{aligned}
\end{equation}

\subsection{Any-variate Causal Attention}\label{sec::attention}
The basic causal attention mechanism constrains each token to focus only on preceding tokens through a causal mask. Prior works typically apply this approach to single-channel data or channel-independent models, overlooking inter-series dependencies. However, correlated variables may reveal shared trends or structural patterns, providing informative context that helps the model more precisely inferring the temporal variations of other variables.
Accordingly, we aim to develop an any-variate causal attention mechanism that incorporates cross-variable information while preserving temporal causality. Following Moirai \citep{moirai}, we flatten the multivariate tokens $H\in\mathbb{R}^{C\times N\times d}$ to $\mathbb{R}^{(C*N)\times d}$ and calculate the attention scores as defined in eq \ref{eq::pos}:

\begin{equation}\label{eq::pos}
\begin{aligned}
\tilde{\mathcal{A}}_{(i-1)N+m,(j-1)N+n}=&(\mathbf{W}_Qh_{i,m})^\top\mathbf{R}_{\Theta,m-n}(\mathbf{W}_Kh_{j,n})\\
&+u\cdot\mathds{1}(i=j)+v\cdot\mathds{1}(i\ne j),\\
i,j\in\{1,2,...,C\}&\text{, }m,n\in\{1,2,...,N\}\text{ ,}
% \mathcal{A}_{ij,mn}=&{\frac{\exp(\tilde{\mathcal{A}}_{ij,mn})}{\sum_{k=1}^{C}\sum_{o=1}^{N}\exp(\tilde{\mathcal{A}}_{ik,mo})}},
\end{aligned}
\end{equation}
where $W_Q,W_K\in\mathbb{R}^{d\times d}$ are the weight matrices, $\mathbf{R}_\Theta$ refers to the rotary matrix\citep{su2024roformer}, $\mathds{1}$ is the indicator function and $u,v$ are two learnable parameters that ensure the permutation equivalence of $\mathcal{A}$ across variables.
We further use a causal mask $\mathbf{M}$ defined in eq \ref{eq::attention} to properly construct the causal relation among tokens within different variables. 
We first calculate the Kronecker product of the adjacency matrix $\mathbf{G}\in\mathbb{R}^{C\times C}$ and the temporal causal mask $\mathbf{T}\in\mathbb{R}^{N\times N}$ and save the results in $\tilde{\mathbf{M}}$, as defined in eq \ref{eq::kron}. $\mathbf{T}$ is a lower triangular mask that masks the attention scores from each token to future positions. The temporal constraint is then broadcast to related variables via the adjacency matrix $\mathbf{G}$. Specifically, for any $h_{i,m}$ and $h_{j,n}$ with $j \ne i$, if $\mathbf{G}_{i,j}=1$, then $h_{i,m}$ and $h_{j,n}$ are mutually dependent for all $n \le m$.

\begin{equation}\label{eq::kron}
\begin{aligned}
\tilde{\mathbf{M}}_{(i-1)N+m,(j-1)N+n}&=\mathbf{G}_{i,j} \mathbf{T}_{m,n}\text{ ,}\\
\mathbf{T}_{m,n}&=\begin{cases}
0&\text{if }m\le n\\
1&\text{otherwise}\text{ ,}
\end{cases}\\
i,j\in\{1,2,...,C\}&\text{, }m,n\in\{1,2,...,N\}\text{ ,}
\end{aligned}
\end{equation}

Subsequently, we construct the mask matrix $\mathbf{M}$ through eq \ref{eq::mask}.
Finally, we apply $\mathbf{M}$ to the attention scores $\mathcal{A}$ to establish causal and variable-wise dependencies among the multivariate tokens, as defined in eq \ref{eq::attention}:

\begin{equation}\label{eq::mask}
\begin{aligned}
\mathbf{M}_{i,j}&=\mathrm{mask}(\tilde{\mathbf{M}}_{i,j})\text{, }i,j\in\{1,2,...,C*N\}\text{,}\\
\mathrm{mask}(x)&=\begin{cases}
-\infty&\text{if }x=0\\
1&\text{otherwise ,}
\end{cases}
\end{aligned}
\end{equation}

\begin{equation}\label{eq::attention}
\begin{aligned}
\mathrm{Attention}(H^{(\ell)})&=\mathrm{Unflat}\left(\mathcal{S}\cdot\mathrm{Flat}(W_VH^{(\ell)})\right)\\
\mathcal{S^{(\ell)}}&=\mathrm{Softmax}(\frac{\mathcal{A}+\mathbf{M}}{\sqrt{d}})
\end{aligned}
\end{equation}
where $W_V\in\mathbb{R}^{d\times d}$ is the weight matrix for the values vectors, $\mathcal{S}^{(\ell)}$ is the attention scores of layer $\ell$, $\ell\in\{1,2,...,J\}$ and $J$ is the number of model layers. Note that $H^{(1)}$ refers to the results of patch-wise tokens, $\mathrm{Flat}$ and $\mathrm{Unflat}$ are the functions that convert the shape of inputs to $\mathbb{R}^{(C*N)\times d}$ and $\mathbb{R}^{C\times N\times d}$ respectively.

\begin{algorithm}[t]
\caption{The implementation of the Time Tracker Encoder with channel-wise MOE}\label{alg::moe}
\renewcommand{\algorithmicrequire}{\textbf{Input:}}
\renewcommand{\algorithmicensure}{\textbf{Output:}}
\begin{algorithmic}[1]
\REQUIRE Input sequence tokens $H^{(\ell)}\in\mathbb{R}^{C\times N\times d}$
\ENSURE Output representations $H^{(\ell+1)}\in\mathbb{R}^{C\times N\times d}$
\setstretch{1.2}
\STATE $\hat{H}^{(\ell)}\in\mathbb{R}^{C\times N\times d}\leftarrow\mathrm{RMSNorm}\left(\mathrm{Attn}(H^{(\ell)})+H^{(\ell)}\right)$
\STATE $s\in\mathbb{R}^{C\times N\times n_p}\leftarrow\hat{H}^{(l)}(\mathcal{G}^{(\ell)})^\top$
\STATE $\bar{s}\in\mathbb{R}^{C\times n_p} \leftarrow \mathrm{Softmax}\left(\frac{1}{N} \sum_{i=1}^{N} s_{:, i, :}\right)$
\STATE Initialize empty vectors $g\in\mathbb{R}^{C\times N\times n_p}, c\in\mathbb{R}^{n_p}$
\FOR{$i=1$ to $C$}
\FOR{$j=1$ to $n_p$}
\IF{$\bar{s}_{i,j}+b_j^{(\ell)}\in \mathrm{TopK}(\{\bar{s}_{i,k}+b_k^{(\ell)}|1\le k\le n_p\},K)$}
\STATE $g_{i,:,j}\leftarrow\bar{s}_{i,j}$
\STATE $c_j\leftarrow c_j+1$
\ELSE
\STATE $g_{i,:,j}\leftarrow0$
\ENDIF
\ENDFOR
\ENDFOR
\STATE $\bar{c}\leftarrow\frac{1}{n_p}\sum_{i=1}^{n_p} c_i$
\FOR{$i=1$ to $n_p$}
\STATE $e=\bar{c}-c_i$
\STATE $b_i^{(\ell)}=b_i^{(\ell)}+u*\mathrm{sign}(e)$
\ENDFOR
\STATE $\tilde{H}_s^{(\ell+1)}\in\mathbb{R}^{C\times N\times d}\leftarrow\frac{1}{n_s}\sum_{i=1}^{n_s}\mathrm{S\_FFN}_i(\hat{H}^{(\ell)})$
\STATE $\tilde{H}_p^{(\ell+1)}\in\mathbb{R}^{C\times N\times d}\leftarrow\sum_{i=1}^{n_p}\mathrm{P\_FFN}_i(\hat{H}^{(\ell)})* g_{:,:,i}$
\STATE $\tilde{H}^{(\ell+1)}\in\mathbb{R}^{C\times N\times d}\leftarrow\tilde{H}_s^{(\ell+1)}+\tilde{H}_p^{(\ell+1)}$
\STATE $H^{(\ell+1)}\in\mathbb{R}^{C\times N\times d}\leftarrow\mathrm{RMSNorm}(\tilde{H}^{(\ell+1)}+\hat{H}^{(\ell)})$
\STATE \textbf{return} $H^{(\ell+1)}$
\end{algorithmic}
\end{algorithm}

\subsection{Channel-wise Mixture of Experts}
Real-world time series often contain diverse and non-stationary patterns across different variables, making it difficult for a single model to generalize well across all conditions\citep{sun2024learning}. In addition, time series data inherently exhibit strict temporal dependencies, where adjacent tokens are often semantically correlated. This motivate us to assign expert networks to tokens within the same univariate sequence to enhance the model's stability and expressiveness. Specifically, we replace the each feedforward network (FFN)\citep{vaswani2017attention} with an MOE layer, containing $n_s$ shared experts and $n_p$ private experts. Each expert network keeps the same architecture of a standard FFN. We design a token cluster $\mathcal{G}\in \mathbb{R}^{n_p\times d}$ to match the tokens within a univariate sequence to the private expert responsible for the corresponding data distribution.
To avoid the issue of routing collapse\citep{shazeer2017outrageously}, we introduce a channel-wise bias factor $b\in\mathbb{R}^{n_p}$ to achieve auxiliary-loss-free load balance\citep{wang2024auxiliary}. The entire computation process of the attention and MOE layer in algorithm \ref{alg::moe}, where the function $\mathrm{Attn}(\cdot)$ corresponds to eq \ref{eq::attention}, $K$ refers to the top-k elements and $\mathrm{S\_FFN, P\_FFN}$ are the shared and private experts.

\section{Experiments}
\subsection{Datasets}
We follow the work of Timer and pretrain our model on Unified Time Series Dataset (UTSD)\citep{liu2024timer}. UTSD is a large time series dataset derived from publicly available online data repositories and real-world machine operation data. It covers seven major domains including energy, environment, health, IoT, nature, transportation, and networks, with up to 1 billion time points. Each subset within UTSD is analyzed in terms of stationarity and predictability to ensure an appropriate level of inherent complexity. To evaluate the generality and adaptivity of our model, we also choose six well-known real-world datasets including Weather, Electricity, Traffic and ETT series. The details of these datasets are shown in table \ref{tab:data}. All these datasets can reached through previous works\citep{wu2021autoformer,itransformer}.

\begin{table}[h]
    \centering
    \scriptsize
    \caption{Statistics of all datasets.}
    \label{tab:data}
    \scalebox{1.0}{
        \begin{tabular}{llll}
        \toprule 
        Dataset & Variables & Frequency & Length \\
        \midrule  
        Weather     & 21  & 10 min & 52696 \\
        Electricity & 321 & 1 hour & 26304 \\
        Traffic     & 862 & 1 hour & 17544 \\
        ETTh1       & 7   & 1 hour & 17420 \\
        ETTh2       & 7   & 1 hour & 17420 \\
        ETTm1       & 7   & 15 min & 69680 \\
        ETTm2       & 7   & 15 min & 69680 \\
        \bottomrule 
        \end{tabular}
    }
\end{table}\begin{table*}[!htbp]
\centering
\large
\caption{Experimental results of long-term multivariate time series forecasting task on 7 real-world datasets. The best results are in \textbf{bold}. The lengths of input and predicting series are $L=96$ and $F=96$ respectively. The modeling types are annotated behind each model, where \textbf{CI} refers to channel-independent and \textbf{CM} refers to channel-mixing. Theses annotations indicate whether the model considers inter-series dependencies.}
\label{tab:result total}
\scriptsize
\renewcommand{\arraystretch}{1.25}
\scalebox{1.0}{
\begin{tabular}{cc|cc|cc|cc|cc|cc|cc|cc}
    \toprule
    \multicolumn{2}{c|}{\multirow{2}{*}{\textbf{Models}}} &
    \multicolumn{2}{c|}{\textbf{Time Tracker}} &
    \multicolumn{2}{c|}{Timer} &
    \multicolumn{2}{c|}{iTransformer} &
    \multicolumn{2}{c|}{PatchTST} &
    \multicolumn{2}{c|}{Autoformer} &
    \multicolumn{2}{c|}{DLinear} &
    \multicolumn{2}{c}{TimesNet} \\
    & & 
    \multicolumn{2}{c|}{\textbf{(CM)}} &
    \multicolumn{2}{c|}{(CI)} & 
    \multicolumn{2}{c|}{(CM)} & 
    \multicolumn{2}{c|}{(CI)} & 
    \multicolumn{2}{c|}{(CM)} & 
    \multicolumn{2}{c|}{(CI)} & 
    \multicolumn{2}{c}{(CI)} \\
    % \multicolumn{2}{c|}{Models}& \multicolumn{2}{c|}{Time Tracker(CM)}& \multicolumn{2}{c|}{Timer}& \multicolumn{2}{c|}{iTransformer} & \multicolumn{2}{c|}{PatchTST} & \multicolumn{2}{c|}{Autoformer} & \multicolumn{2}{c|}{DLinear} & \multicolumn{2}{c}{TimesNet}\\
    \cmidrule(lr){1-2}\cmidrule(lr){3-4}\cmidrule(lr){5-6}\cmidrule(lr){7-8}\cmidrule(lr){9-10}\cmidrule(lr){11-12}\cmidrule(lr){13-14}\cmidrule(lr){15-16}
    \multicolumn{2}{c|}{Metric}&MSE&MAE&MSE&MAE&MSE&MAE&MSE&MAE&MSE&MAE&MSE&MAE&MSE&MAE\\
    \midrule
    \multicolumn{2}{c|}{Weather} & \textbf{0.169} & \textbf{0.213} & 0.176 & 0.217 & 0.175 & 0.216 & 0.172 & 0.214 & 0.249 & 0.329 & 0.198 & 0.260 & 0.172 & 0.220 \\
    \midrule
    \multicolumn{2}{c|}{Traffic} & \textbf{0.370} & \textbf{0.249} & 0.407 & 0.260 & 0.392 & 0.263 & 0.444 & 0.283 & 0.597 & 0.371 & 0.652 & 0.386 & 0.593 & 0.321 \\
    \midrule
    \multicolumn{2}{c|}{Electricity} & \textbf{0.141} & \textbf{0.233} & 0.158 & 0.242 & 0.153 & 0.245 & 0.167 & 0.253 & 0.196 & 0.313 & 0.195 & 0.278 & 0.168 & 0.272 \\
    \midrule
    \multicolumn{2}{c|}{ETTh1} & 0.380 & 0.400 & \textbf{0.379} & \textbf{0.399} & 0.394 & 0.410 & 0.380 & \textbf{0.399} & 0.435 & 0.446 & 0.391 & 0.403 & 0.384 & 0.402 \\
    \midrule
    \multicolumn{2}{c|}{ETTh2} & 0.294 & 0.347 & 0.309 & 0.356 & 0.303 & 0.353 & \textbf{0.293} & \textbf{0.342} & 0.332 & 0.368 & 0.375 & 0.397 & 0.340 & 0.374 \\
    \midrule
    \multicolumn{2}{c|}{ETTm1} & \textbf{0.325} & \textbf{0.363} & 0.330 & 0.367 & 0.339 & 0.374 & 0.326 & \textbf{0.363} & 0.510 & 0.492 & 0.344 & 0.372 & 0.338 & 0.375 \\
    \midrule
    \multicolumn{2}{c|}{ETTm2} & 0.178 & \textbf{0.261} & 0.180 & 0.201 & 0.189 & 0.274 & \textbf{0.177} & \textbf{0.261} & 0.205 & 0.293 & 0.190 & 0.287 & 0.187 & 0.267 \\
    \bottomrule
\end{tabular}
}
\end{table*}\begin{table*}[!htbp]
\centering
\large
\caption{Zero-shot results of long-term multivariate time series forecasting task on 7 real-world datasets. The best results are in \textbf{bold}. Time Tracker is pretrained on UTSD dataset, which is kept same for pretraining Time-XL and Timer. The lengths of input and predicting series are $L=672$ and $F=96$ respectively.}
\label{tab:zero}
\scriptsize
\renewcommand{\arraystretch}{1.25}
\scalebox{1.0}{
\begin{tabular}{cc|cc|cc|cc|cc|cc}
    \toprule
    \multicolumn{2}{c|}{\multirow{2}{*}{\textbf{Models}}} &
    \multicolumn{2}{c|}{$\textbf{Time Tracker}_{\textit{Base}}$} &
    \multicolumn{2}{c|}{$\textbf{Timer}_{\textit{Base}}$} &
    \multicolumn{2}{c|}{$\textbf{Moirai}_{\textit{Base}}$} &
    \multicolumn{2}{c|}{$\textbf{MOMENT}$} &
    \multicolumn{2}{c}{$\textbf{Chronos}_{\textit{Base}}$} \\
    & & 
    \multicolumn{2}{c|}{\textbf{(Ours)}} &
    \multicolumn{2}{c|}{(\citeyear{liu2024timer})} & 
    \multicolumn{2}{c|}{(\citeyear{moirai})} & 
    \multicolumn{2}{c|}{(\citeyear{goswami2024moment})} & 
    \multicolumn{2}{c}{(\citeyear{ansarichronos})} \\
    \cmidrule(lr){1-2}\cmidrule(lr){3-4}\cmidrule(lr){5-6}\cmidrule(lr){7-8}\cmidrule(lr){9-10}\cmidrule(lr){11-12}
    \multicolumn{2}{c|}{Metric}&MSE&MAE&MSE&MAE&MSE&MAE&MSE&MAE&MSE&MAE\\
    \midrule
    \multicolumn{2}{c|}{Weather} & 0.173 & \textbf{0.221} & \textbf{0.172} & 0.224 & 0.175 & 0.210 & 0.243 & 0.255 & 0.203 & 0.238  \\
    \midrule
    \multicolumn{2}{c|}{Traffic} & \textbf{0.413} & \textbf{0.286} & 0.471 & 0.322 & 0.449 & 0.306 & 0.507 & 0.331 & 0.440 & 0.299 \\
    \midrule
    \multicolumn{2}{c|}{Electricity} & \textbf{0.150} & 0.241 & 0.165 & 0.257 & 0.201 & 0.278 & 0.291 & 0.355 & 0.153 & \textbf{0.231} \\
    \midrule
    \multicolumn{2}{c|}{ETTh1} & \textbf{0.376} & 0.400 & 0.388 & 0.406 & 0.392 & 0.402 & 0.688 & 0.557 & 0.440 & \textbf{0.393} \\
    \midrule
    \multicolumn{2}{c|}{ETTh2} & 0.303 & 0.354 & 0.305 & 0.355 & \textbf{0.284} & \textbf{0.331} & 0.342 & 0.396 & 0.308 & \textbf{0.343}  \\
    \midrule
    \multicolumn{2}{c|}{ETTm1} & \textbf{0.386} & \textbf{0.396} & 0.552 & 0.473 & 0.447 & 0.403 & 0.654 & 0.527 & 0.454 & 0.408 \\
    \midrule
    \multicolumn{2}{c|}{ETTm2} & \textbf{0.189} & \textbf{0.273} & 0.222 & 0.294 & 0.219 & 0.290 & 0.260 & 0.335 & 0.199 & 0.274 \\
    \bottomrule
\end{tabular}
}
\end{table*}

\begin{table*}[!htbp]
\centering
\large
\caption{Few-shot results of long-term multivariate time series forecasting task on 7 real-world datasets. The best results are in \textbf{bold}. The models are all pretrained on UTSD dataset, while Time Tracker adopts multivariate finetuning and Timer uses univariate finetuning. The lengths of input and predicting series are $L=672$ and $F=96$ respectively.}
\label{tab:few}
\scriptsize
\renewcommand{\arraystretch}{1.25}
\scalebox{1.0}{
\begin{tabular}{cc|cc|cc|cc|cc|cc|cc|cc}
    \toprule
    \multicolumn{2}{c|}{Datasets} &
    \multicolumn{2}{c|}{Weather} &
    \multicolumn{2}{c|}{Traffic} &
    \multicolumn{2}{c|}{Electricity} &
    \multicolumn{2}{c|}{ETTh1} &
    \multicolumn{2}{c|}{ETTh2} &
    \multicolumn{2}{c|}{ETTm1} &
    \multicolumn{2}{c}{ETTm2} \\
    \cmidrule(lr){1-2}\cmidrule(lr){3-4}\cmidrule(lr){5-6}\cmidrule(lr){7-8}\cmidrule(lr){9-10}\cmidrule(lr){11-12}\cmidrule(lr){13-14}\cmidrule(lr){15-16}
    \multicolumn{2}{c|}{Metric}&MSE&MAE&MSE&MAE&MSE&MAE&MSE&MAE&MSE&MAE&MSE&MAE&MSE&MAE\\
    \midrule
    \multicolumn{2}{c|}{Time Tracker} & 0.154 & 0.200 & \textbf{0.353} & \textbf{0.244} & \textbf{0.128} & \textbf{0.221} & \textbf{0.364} & \textbf{0.391} & \textbf{0.294} & \textbf{0.351} & \textbf{0.341} & \textbf{0.369} & \textbf{0.171} & \textbf{0.254} \\
    \midrule
    \multicolumn{2}{c|}{Timer} & \textbf{0.151} & \textbf{0.198} & 0.362 & 0.247 & 0.132 & 0.225 & 0.378 & 0.398 & 0.296 & 0.356 & 0.395 & 0.393 & 0.177 & 0.256 \\
    \bottomrule
\end{tabular}
}
\end{table*}

\subsection{Model Parameter Settings}\label{sec::settings}
By default, we set the input and output lengths to $L = 672$ and $F = 96$, respectively. Each patch is projected into a token with embedding dimension $d = 512$. To improve computational efficiency while preserving model capacity, we adopt a sparse placement strategy for the channel-wise MOE modules, where Dense MLP layers and MoE layers are alternately inserted after each Attention layer. Specifically, each Dense MLP expands the token representation by a factor of 4 in the hidden dimension, following a standard feed-forward design. In contrast, within the MoE layers, we reduce the dimensional expansion of each private expert by a factor of $\frac{4}{K}$, which helps maintain a comparable overall computational budget while enabling multiple experts to be selectively activated.

\subsection{Multivariate Forecasting}
\subsubsection{Setup}
To evaluate the effectiveness of the proposed model structure, we follow iTransformer\citep{itransformer} and conduct experiments of multivariate forecasting on the six benchmark datasets. The lengths of training, validation and test sets are recorded in table \ref{tab:data}. We set $L=96$ and $F=96$ and present the results of MSE and MAE in table \ref{tab:result total}. Since we aim to directly predict multivariate series, we activate the graph learning layer and use the adjacency matrix $\mathbf{G}$ in all $J$ layers.
For Weather, Traffic and Electricity that have more variables and exhibit better seasonality, we set $J=4$ to capture more complex inter-series dependencies. For ETT datasets with relatively weak relationships between variables, we set $J=2$ to slightly weaken the emphasis on inter-series dependencies.
\subsubsection{Results}
Generally, Time Tracker achieves advanced predicting performance on all the datasets among both channel-mixing models and channel independent models. Compared to iTransformer, the SOTA channel-mixing model, Time Tracker reduces the MSE and MAE by 4.42\% and 3.23\% on average. When dealing with datasets with weak inter-series relationships, Time Tracker can also reach or outperform the SOTA channel-independent models, including a LTSM benchmark model Timer. This indicates that the structure of Time Tracker can effectively capture both temporal dependencies within the sequence and inter-series dependencies, regardless of whether the data exhibits high or low variable correlations.

\subsection{Zero-shot Learning}\label{sec::zero}
\subsubsection{Setup}
In this section, we pretrain Time Tracker in the univariate manner on UTSD and conduct experiments on six well-known datasets. For all the baseline models, we set the input size as $L=672$ and patch size as $P=96$. We record the MSE and MAE results of $F=96$ in table \ref{tab:zero}. Note that for fair comparison, we follow the model structure of other baseline models and pretrain them with $L=672$. We use the pretrained version of baseline models to get zero-shot results.
% For prediction lengths that exceed the patch size, we perform rolling forecasting, using the previous prediction results as the input sequence to autoregressively generate longer forecasts.
\subsubsection{Results}
Time Tracker achieves superior predicting performance. Compared to the SOTA multivariate pretrained model Moirai, the MSE and MAE are reduced by 9.59\% and 2.79\%. This improvement comes to 10.98\% and 6.77\% relative to the results obtained by Timer, which is pretrained in a univariate manner. Generally, pretraining models in a univariate way could help enhancing the model's generality. 
What's more, we believe that the MOE layers can better generalize across diverse temporal patterns  by assigning specific experts to different types of time series. In the absence of MoE, FFN layers have to fit all data distributions simultaneously, which may results in suboptimal network parameters when dealing with newly seen data. By contrast, MoE enables experts to specialize in prototype-like data distribution. Each expert adapts to a specific distribution in the high-dimensional space. When dealing with data within new distributions, MOE enables the model to activate parameters closer to the target pattern rather than relying on the averaged ones across all pretrained data.\subsection{Few-shot Learning}
\subsubsection{Setup}
In this section, to evaluate the model's adaptability, we conduct few-shot learning experiments on six datasets. Different from the pretraining stage, we introduce the graph learning layer to generate an adjacency matrix $\mathbf{G}$ to guide the interaction among tokens from different variables. We set the first $\ell_u=7$ Time Tracker layers remaining the parameters to process the input data in a channel-independent way. The last $\ell_m=1$ Time Tracker layer will concatenate all sequences to the multivariate form and capture inter-series dependencies. Since the parameters of the attention layers and normalization layers are independent of the number of tokens, all parameters from the pretrained model will be kept to enable faster convergence. We use Timer as the benchmark model and conduct univariate finetuning. All the time series will be input independently to the model.
We use only 20\% of the dataset as the training set while 20\% as the testing set and record the MSE and MAE results in table \ref{tab:few}.
\subsubsection{Results}
Compared to the univariately finetuned model Timer, Time Tracker demonstrates superior adaptability to complex datasets, particularly those involving multiple variables with stronger interdependencies. The average improvements of MSE and MAE are 3.56\% and 2.34\%. Our proposed graph learning module adaptively constructs inter-series relationships for previously unseen multivariate data and integrates the relational information into the any-variate attention layer. Compared to maintaining a univariate finetuning strategy, our design facilitates propagating the temporal patterns across sequence tokens from variables with strong correlations. With the help of inter-series dependencies, Time Tracker exhibits better adaptability to specific datasets.

\subsection{Ablation Studies}
\subsubsection{Token-wise and Channel-wise MOE Routing Strategies}
By default, all of the patch-wise tokens generated from the same input series will be assigned with the same expert network. Such design is notably different from the routing strategy commonly adopted in conventional MOE architectures\citep{shi2024timemoe}, where expert selection is typically performed at the token level. To explicitly examine the effect of routing granularity, we conduct an ablation study by replacing the proposed channel-wise MoE routing with a token-wise routing strategy. Specifically, under the token-wise setting, each token independently selects its expert according to the gating function, while keeping all other architectural components unchanged. For a fair comparison, MoE layers are retained in the same transformer blocks in both settings. All models are pretrained on the same UTSD dataset under identical training configurations and their performance is evaluated on the same samples under zero-shot scenarios. This experimental design allows us to isolate the impact of different MoE routing strategies on downstream time series prediction performance.

\begin{figure}
\centering
\includegraphics[width=1.0\linewidth]{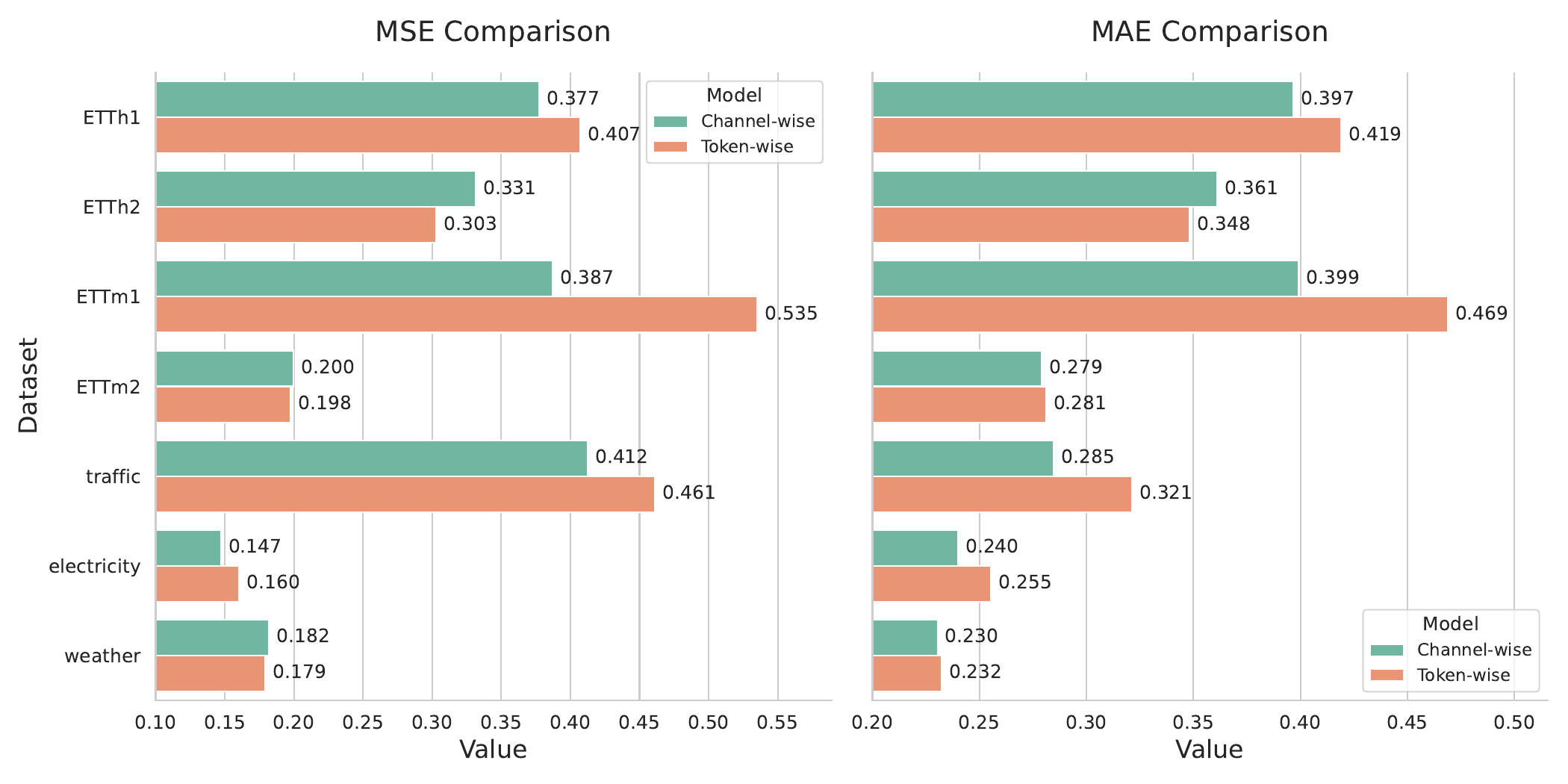}
\caption{The MSE and MAE scores of Time Tracker using channel-wise MOE and token-wise MOE routing strategies. Results are reported under a zero-shot setting.}
\label{fig:channel_token}
\end{figure}

As shown in Fig \ref{fig:channel_token}, the MSE and MAE scores of channel-wise MOE setting is generally lower than that of the token-wise MOE setting. The observed performance gap between token-wise and channel-wise MoE routing can be attributed to the intrinsic characteristics of time series data.
Unlike natural language tokens, patch-wise tokens extracted from the same time series channel are not semantically independent. Instead, they typically share common generative factors such as trends, seasonal patterns and noise characteristics. Token-wise routing implicitly assumes that each token may benefit from a distinct expert, which can be an overly strong assumption in time series modeling and may disrupt temporal coherence by assigning adjacent tokens from the same variable to different experts. In contrast, channel-wise routing enforces a consistent expert assignment across all tokens belonging to the same input series, enabling each expert to capture complete temporal dynamics at the series level. This design provides a stronger inductive bias aligned with the underlying data structure, improves routing stability, and reduces the influence of local fluctuations or noise on the gating decision.
As a result, channel-wise routing encourages clearer expert specialization with respect to series-level statistical properties, which may lead to more robust representations and better generalization performance, particularly in zero-shot forecasting scenarios.

\begin{figure}
\centering
\subfigure[Results of MSE on different datasets.]{
    \includegraphics[width=0.9\linewidth]{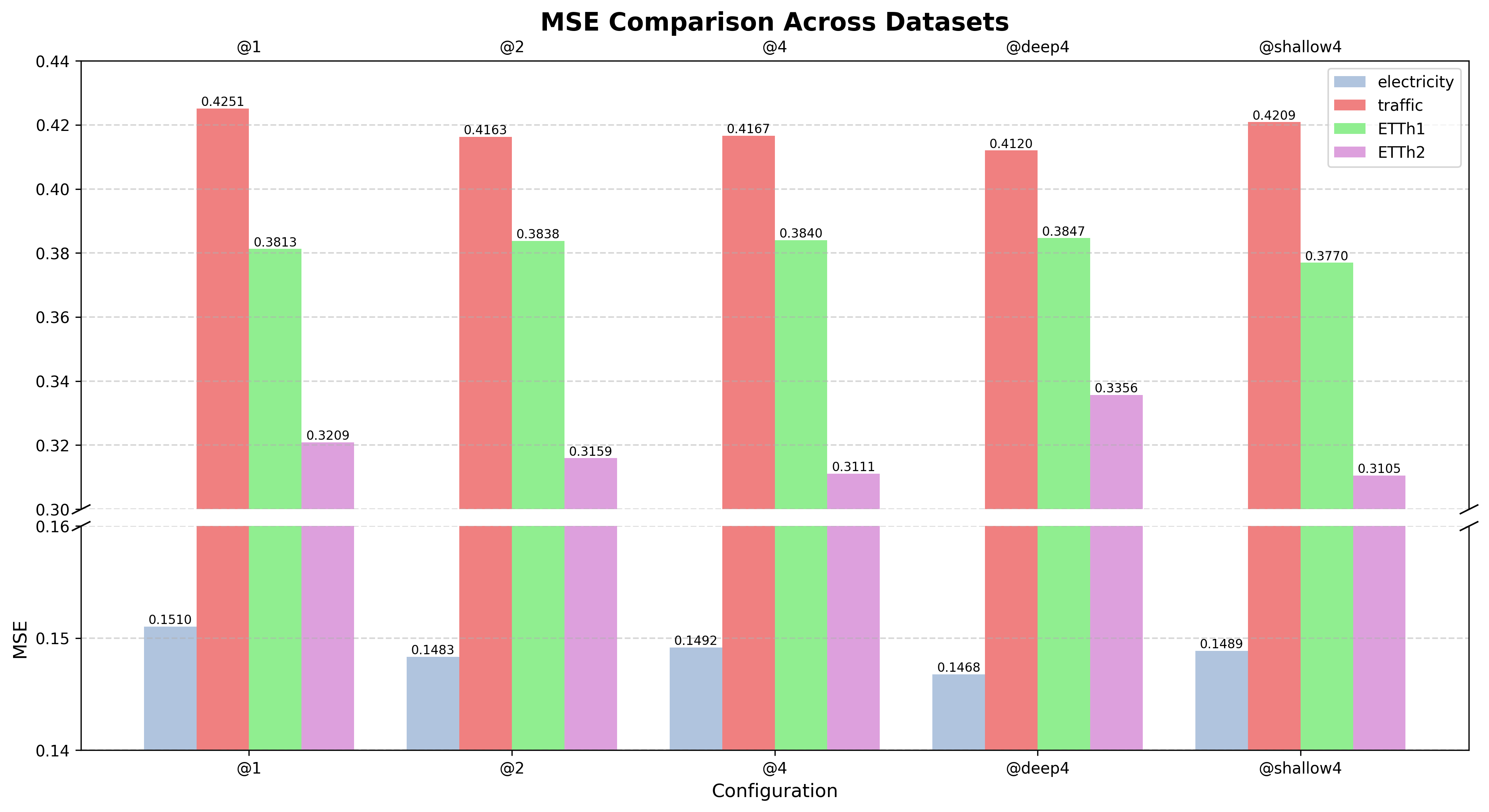}
}
\subfigure[Results of $\mathrm{R}^2$ on different datasets.]{
    \includegraphics[width=0.9\linewidth]{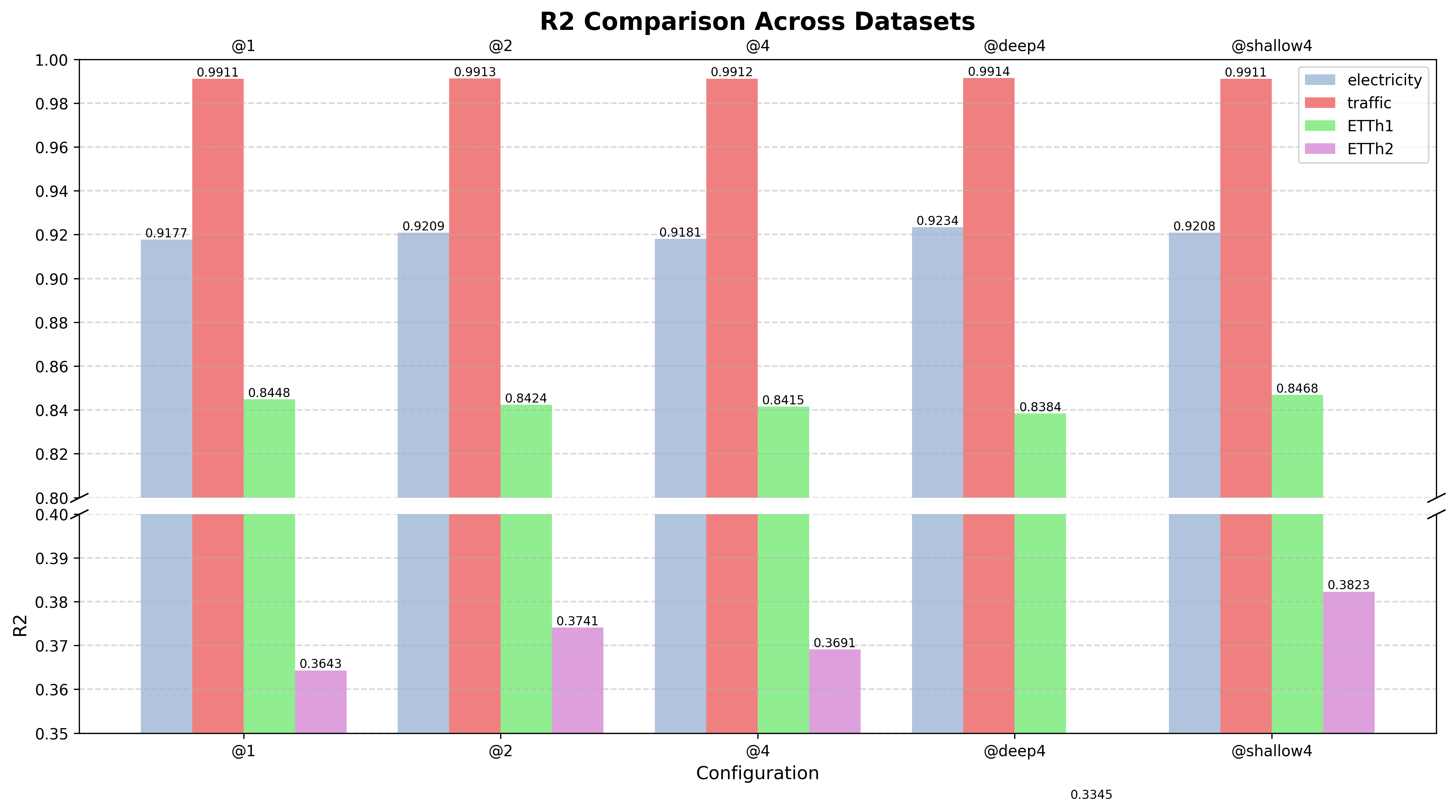}
}
\caption{MSE and $\mathrm{R}^2$ of varying expert density settings on Electricity, Traffic, ETTh1 and ETTh2.}
\label{fig:moe density}
\end{figure}

\subsubsection{Impact of MOE Layer Density}
In TimeTracker, different expert networks share the same feed-forward architecture and apply nonlinear transformations to the outputs of the attention layers, thereby enhancing the expressive capacity of the model. By default, each decoder layer incorporates an MoE module, enabling experts to process tokens associated with different representation distributions. We further explore the representations processed by MoE layers across network depth. Since MoE modules operate by conditionally transforming layer-wise representations, their contribution may depend on the nature of the underlying attention-induced features. Motivated by this, we conduct experiments by selectively adjusting the density of MOE layers across network depth. We make the following settings: (a) applying MOE at all layers; (b) inserting MOE every 2 layers; (c) inserting MOE every 4 layers; (d) preferentially applying MOE at shallow layers; (e) preferentially applying MOE at deep layers. Considering that Time Tracker has 8 decoder layers by default, we initialize 4 MoE layers and embed them into either the first 4 layers for setting (d) or the last 4 layers for setting (e).

Given the above settings, we re-pretrain a set of model parameters accordingly. We then perform zero-shot experiments on Traffic, Electricity, ETTh1, and ETTh2 following the data loading strategy described in Section \ref{sec::zero}. We display the MSE and $\mathrm{R}^2$ in Figure \ref{fig:moe density}. It can be observed that distributing MoE across every model layer is not necessarily the optimal solution. Instead, sparsely placing MoE modules at selected layers can simultaneously improve predictive performance and computational efficiency. For datasets with clearer periodic patterns and more stable temporal dynamics, such as Traffic and Electricity, placing MoE modules in the deeper layers of the model tends to yield greater performance gains. In contrast, for ETT datasets with weaker periodicity and less stable temporal patterns, placing MoE modules in the shallower layers leads to better forecasting results. This phenomenon can be further interpreted from the perspective of hierarchical representation learning in Transformer-based models. In general, the shallow layers of a Transformer tend to capture low-level temporal features, such as short-term fluctuations, local dependencies and basic statistical patterns of the time series. These representations are relatively close to the raw input and are sensitive to local variations. On the other hand, deeper layers gradually integrate information across longer temporal contexts and multiple variables, forming higher-level abstractions that reflect more global structures, such as long-term trends, periodic patterns, and stable inter-series relationships.

Based on this hierarchical learning behavior, the optimal placement of MoE modules may depend on the intrinsic characteristics of the dataset. When the time series exhibits strong periodicity and stable long-term patterns, deeper layers play a greater role in modeling these structured global dependencies. Introducing MoE at these layers allows different experts to specialize in distinct long-term patterns or periodic structures, thereby improving the model’s ability to capture subtle variations across cycles. Conversely, when the data contains irregular fluctuations and lacks clear periodic patterns, the predictive signal is often dominated by short-term dynamics and local variations. In such cases, allocating MoE modules to earlier layers enables experts to specialize in different local behaviors or short-term patterns, which can improve the model's flexibility in fitting unstable temporal dynamics.

This observation suggests a potentially useful design principle for MoE-based time series forecasting models: rather than uniformly inserting experts across all layers, the placement of MoE modules should be adapted to the structural characteristics of the dataset. More broadly, it indicates that expert specialization in time series models may align with the hierarchical feature extraction process of deep architectures, where different layers correspond to different levels of temporal patterns. Designing expert routing strategies that are aware of this hierarchy may further enhance both model efficiency and forecasting accuracy.

% \subsection{Impact of multivariate finetuning mode}

\begin{figure}
    \centering
    \includegraphics[width=1.0\linewidth]{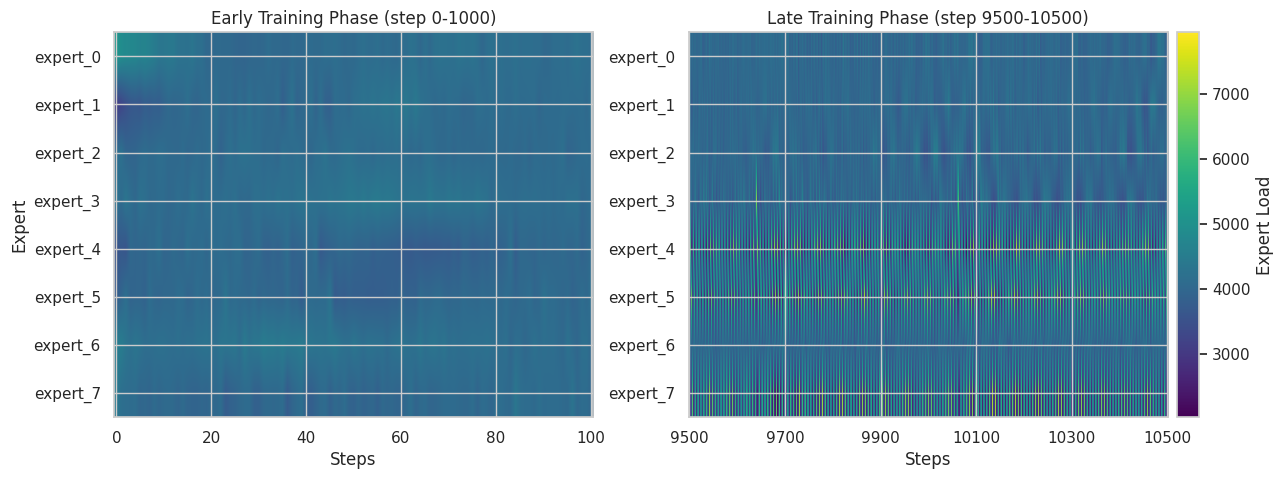}
    \caption{Expert load distribution during training iterations. The utilization of different experts at both the early stage (0–100 iterations) and the late stage (9500–10500 iterations) is relatively stable.}
    \label{fig:loading balance}
\end{figure}

\subsection{Load balancing of Experts}
In the standard Mixture of Experts architecture, expert routing is typically performed at the token level, where each token independently selects its expert through a gating network. This fine-grained routing mechanism naturally facilitates balanced expert utilization due to the large number of routing decisions during training.

In contrast, our model adopts a sequence-level expert routing strategy, where all tokens belonging to the same variable are assigned to the same expert. This design may introduces a potential risk of expert load imbalance. Since routing decisions are made at a coarser granularity, the number of assignments is significantly reduced compared to token-level routing, which may lead to certain experts being over-utilized while others remain underused.

To investigate whether the proposed routing strategy affects expert utilization, we analyze the distribution of expert loads at different stages of training. Specifically, we report the expert usage statistics during the early training stage (0–100 iterations) and the later training stage (9500–10500 iterations). As shown in Figure \ref{fig:loading balance}, expert loads remain consistently balanced across all experts in both stages. During the early training phase, the gating network already distributes sequences relatively evenly among experts, indicating stable exploration of different experts. More importantly, the balanced utilization persists in the later training stage, suggesting that the model does not converge to a degenerate routing pattern where only a subset of experts dominates the computation. These observations demonstrate that the proposed sequence-level routing mechanism maintains stable expert utilization throughout training and does not suffer from expert collapse. Moreover, the results indicate that the adopted load-balancing strategy is effective in preventing routing bias and ensuring that all experts are consistently utilized during training. Consequently, the model preserves the key advantage of MoE architectures—efficient parameter utilization through sparse expert activation—while enforcing consistent expert assignment within each variable sequence.

\subsection{Model Parameter Size Analysis}
During the pretraining stage, the parameters of Time Tracker primarily originate from the following components: (a) Patch Embedding Layer, (b) variable identifier embeddings, (c) Transformer Blocks and (c) Output Head. Specifically, each Transformer Block comprises the self-attention layer, Feed-Forward Networks (FFN) (which may be implemented as MoE) and normalization layers. In the fine-tuning stage, a (d) Graph Learning Layer is additionally introduced, incorporating learnable parameters $\alpha$.
\paragraph{Patch Embedding Layer} $\mathbf{W}_p \in \mathbb{R}^{P\times d}$ to map input patches to tokens.
\paragraph{Variable Identifier Embeddings} $\mathbb{E}_{id}\in\mathbb{R}^{2\times h}$ to distinguish whether tokens belong to the same variable. The identifier embeddings are set separate for each attention head.
\paragraph{Self-Attention Layer} $\mathbf{W}_q\in\mathbb{R}^{d\times d_q}$, $\mathbf{W}_k, \mathbf{W}_v \in \mathbb{R}^{d\times d_k}$ of $h$ heads and $\mathbf{W}_o \in \mathbb{R}^{d\times d}$ for all heads. By default, $d_q,d_k=d$.
\paragraph{Normalization Layer} We use Layer Normalization (LN) by default. A standard LN layer contains two sets of learnable parameters: (a) Gain ($\gamma \in \mathbb{R}^d$): a scaling vector that re-scales the normalized output and (b) Bias ($\beta \in \mathbb{R}^d$): a shifting vector that offsets the normalized output. In each Transformer Block, there are typically two LN layers right after the Self-Attention Layer and FFN.
\paragraph{Feed-Forward Network} There are basically two linear projectors within a standard FFN: $\mathbf{W}_{\text{ffn1}}, \mathbf{W}_{\text{ffn2}} \in  \mathbb{R}^{d\times d_{\text{ff}}}$. If we use MOE experts, an additional gating is introduced where $\mathbf{W}_g\in\mathbb{R}^{d\times d_{\text{ff}}}$.
\paragraph{Output Head} $\mathbf{W}_{d} \in \mathbb{R}^{d\times P}$ to map the tokens to output patches.
\paragraph{Graph Learning Layer} $\alpha\in\mathbb{R}^{\frac{L}{2}}$ and three learnable edge weight biases $\mathbb{R}$.

According to the parameter settings in Section \ref{sec::settings}, the total activated parameters during pretraining and fine-tuning stages are 79,911,648 and 16,850,883 respectively.

\section{Conclusion}
In this paper, we propose Time Tracker, a large time series model architecture for multivariate time series forecasting. We seamlessly integrate MOE with Transformer to assign different expert networks to sequence tokens from time series with varying data distributions, aiming to produce higher-quality features. We design an any-variate causal attention mechanism that can process tokens from any number of variables with the same model structure. Such a module allows for different data loading strategies in the pretraining and finetuning stages. We pretrain the model under a univariate setting and seamlessly transfer the obtained parameters to the multivariate finetuning stage. The decoupled design effectively enhances the model’s generalization to unseen data distributions and adaptability to specific datasets. Future research will focus on incorporating additional modalities and constructing multimodal time series datasets.

% \bibliography{main}

\begin{thebibliography}{29}
\providecommand{\natexlab}[1]{#1}
\providecommand{\url}[1]{\texttt{#1}}
\expandafter\ifx\csname urlstyle\endcsname\relax
  \providecommand{\doi}[1]{doi: #1}\else
  \providecommand{\doi}{doi: \begingroup \urlstyle{rm}\Url}\fi

\bibitem[Grattafiori et~al.(2024)Grattafiori, Dubey, Jauhri, Pandey, Kadian, Al-Dahle, Letman, Mathur, Schelten, Vaughan, et~al.]{grattafiori2024llama}
Aaron Grattafiori, Abhimanyu Dubey, Abhinav Jauhri, Abhinav Pandey, Abhishek Kadian, Ahmad Al-Dahle, Aiesha Letman, Akhil Mathur, Alan Schelten, Alex Vaughan, et~al.
\newblock The llama 3 herd of models.
\newblock \emph{arXiv preprint arXiv:2407.21783}, 2024.

\bibitem[Liu et~al.(2021)Liu, Lin, Cao, Hu, Wei, Zhang, Lin, and Guo]{liu2021swin}
Ze~Liu, Yutong Lin, Yue Cao, Han Hu, Yixuan Wei, Zheng Zhang, Stephen Lin, and Baining Guo.
\newblock Swin transformer: Hierarchical vision transformer using shifted windows.
\newblock In \emph{Proceedings of the IEEE/CVF international conference on computer vision}, pages 10012--10022, 2021.

\bibitem[Kirillov et~al.(2023)Kirillov, Mintun, Ravi, Mao, Rolland, Gustafson, Xiao, Whitehead, Berg, Lo, et~al.]{kirillov2023segment}
Alexander Kirillov, Eric Mintun, Nikhila Ravi, Hanzi Mao, Chloe Rolland, Laura Gustafson, Tete Xiao, Spencer Whitehead, Alexander~C Berg, Wan-Yen Lo, et~al.
\newblock Segment anything.
\newblock In \emph{Proceedings of the IEEE/CVF international conference on computer vision}, pages 4015--4026, 2023.

\bibitem[Devlin et~al.(2019)Devlin, Chang, Lee, and Toutanova]{devlin2019bert}
Jacob Devlin, Ming-Wei Chang, Kenton Lee, and Kristina Toutanova.
\newblock Bert: Pre-training of deep bidirectional transformers for language understanding.
\newblock In \emph{Proceedings of the 2019 conference of the North American chapter of the association for computational linguistics: human language technologies, volume 1 (long and short papers)}, pages 4171--4186, 2019.

\bibitem[Wu et~al.(2021)Wu, Xu, Wang, and Long]{wu2021autoformer}
Haixu Wu, Jiehui Xu, Jianmin Wang, and Mingsheng Long.
\newblock Autoformer: Decomposition transformers with auto-correlation for long-term series forecasting.
\newblock \emph{Advances in neural information processing systems}, 34:\penalty0 22419--22430, 2021.

\bibitem[Zeng et~al.(2023)Zeng, Chen, Zhang, and Xu]{zeng2023transformers}
Ailing Zeng, Muxi Chen, Lei Zhang, and Qiang Xu.
\newblock Are transformers effective for time series forecasting?
\newblock In \emph{Proceedings of the AAAI conference on artificial intelligence}, volume~37, pages 11121--11128, 2023.

\bibitem[Nie et~al.(2023)Nie, Nguyen, Sinthong, and Kalagnanam]{patchtst}
Yuqi Nie, Nam~H. Nguyen, Phanwadee Sinthong, and Jayant Kalagnanam.
\newblock A time series is worth 64 words: Long-term forecasting with transformers.
\newblock In \emph{The Eleventh International Conference on Learning Representations, {ICLR} 2023, Kigali, Rwanda, May 1-5, 2023}. OpenReview.net, 2023.

\bibitem[Liu et~al.(2024{\natexlab{a}})Liu, Hu, Zhang, Wu, Wang, Ma, and Long]{itransformer}
Yong Liu, Tengge Hu, Haoran Zhang, Haixu Wu, Shiyu Wang, Lintao Ma, and Mingsheng Long.
\newblock itransformer: Inverted transformers are effective for time series forecasting.
\newblock In \emph{The Twelfth International Conference on Learning Representations, {ICLR} 2024, Vienna, Austria, May 7-11, 2024}. OpenReview.net, 2024{\natexlab{a}}.

\bibitem[Liu et~al.(2024{\natexlab{b}})Liu, Zhang, Li, Huang, Wang, and Long]{liu2024timer}
Yong Liu, Haoran Zhang, Chenyu Li, Xiangdong Huang, Jianmin Wang, and Mingsheng Long.
\newblock Timer: generative pre-trained transformers are large time series models.
\newblock In \emph{Proceedings of the 41st International Conference on Machine Learning}, pages 32369--32399, 2024{\natexlab{b}}.

\bibitem[Liu et~al.(2024{\natexlab{c}})Liu, Qin, Huang, Wang, and Long]{liu2024timerxl}
Yong Liu, Guo Qin, Xiangdong Huang, Jianmin Wang, and Mingsheng Long.
\newblock Timer-xl: Long-context transformers for unified time series forecasting.
\newblock \emph{arXiv preprint arXiv:2410.04803}, 2024{\natexlab{c}}.

\bibitem[Chen(2024)]{chen2024model}
Pin-Yu Chen.
\newblock Model reprogramming: Resource-efficient cross-domain machine learning.
\newblock In \emph{Proceedings of the AAAI Conference on Artificial Intelligence}, volume~38, pages 22584--22591, 2024.

\bibitem[Jin et~al.(2024)Jin, Wang, Ma, Chu, Zhang, Shi, Chen, Liang, Li, Pan, et~al.]{jintime}
Ming Jin, Shiyu Wang, Lintao Ma, Zhixuan Chu, James~Y Zhang, Xiaoming Shi, Pin-Yu Chen, Yuxuan Liang, Yuan-Fang Li, Shirui Pan, et~al.
\newblock Time-llm: Time series forecasting by reprogramming large language models.
\newblock In \emph{The Twelfth International Conference on Learning Representations}, 2024.

\bibitem[Rasul et~al.(2023)Rasul, Ashok, Williams, Ghonia, Bhagwatkar, Khorasani, Bayazi, Adamopoulos, Riachi, Hassen, et~al.]{rasul2023lag}
Kashif Rasul, Arjun Ashok, Andrew~Robert Williams, Hena Ghonia, Rishika Bhagwatkar, Arian Khorasani, Mohammad Javad~Darvishi Bayazi, George Adamopoulos, Roland Riachi, Nadhir Hassen, et~al.
\newblock Lag-llama: Towards foundation models for probabilistic time series forecasting.
\newblock \emph{arXiv preprint arXiv:2310.08278}, 2023.

\bibitem[Ansari et~al.(2024)Ansari, Stella, Turkmen, Zhang, Mercado, Shen, Shchur, Rangapuram, Arango, Kapoor, et~al.]{ansarichronos}
Abdul~Fatir Ansari, Lorenzo Stella, Ali~Caner Turkmen, Xiyuan Zhang, Pedro Mercado, Huibin Shen, Oleksandr Shchur, Syama~Sundar Rangapuram, Sebastian~Pineda Arango, Shubham Kapoor, et~al.
\newblock Chronos: Learning the language of time series.
\newblock \emph{Transactions on Machine Learning Research}, 2024.

\bibitem[Goswami et~al.(2024)Goswami, Szafer, Choudhry, Cai, Li, and Dubrawski]{goswami2024moment}
Mononito Goswami, Konrad Szafer, Arjun Choudhry, Yifu Cai, Shuo Li, and Artur Dubrawski.
\newblock Moment: a family of open time-series foundation models.
\newblock In \emph{Proceedings of the 41st International Conference on Machine Learning}, pages 16115--16152, 2024.

\bibitem[Das et~al.(2024)Das, Kong, Sen, and Zhou]{das2024decoder}
Abhimanyu Das, Weihao Kong, Rajat Sen, and Yichen Zhou.
\newblock A decoder-only foundation model for time-series forecasting.
\newblock In \emph{Forty-first International Conference on Machine Learning}, 2024.

\bibitem[Woo et~al.(2024)Woo, Liu, Kumar, Xiong, Savarese, and Sahoo]{moirai}
Gerald Woo, Chenghao Liu, Akshat Kumar, Caiming Xiong, Silvio Savarese, and Doyen Sahoo.
\newblock Unified training of universal time series forecasting transformers.
\newblock In \emph{Forty-first International Conference on Machine Learning, {ICML} 2024, Vienna, Austria, July 21-27, 2024}. OpenReview.net, 2024.

\bibitem[Fedus et~al.(2022)Fedus, Zoph, and Shazeer]{fedus2022switch}
William Fedus, Barret Zoph, and Noam Shazeer.
\newblock Switch transformers: Scaling to trillion parameter models with simple and efficient sparsity.
\newblock \emph{Journal of Machine Learning Research}, 23\penalty0 (120):\penalty0 1--39, 2022.

\bibitem[Liu et~al.(2024{\natexlab{d}})Liu, Liu, Woo, Aksu, Liang, Zimmermann, Liu, Savarese, Xiong, and Sahoo]{liu2024moirai}
Xu~Liu, Juncheng Liu, Gerald Woo, Taha Aksu, Yuxuan Liang, Roger Zimmermann, Chenghao Liu, Silvio Savarese, Caiming Xiong, and Doyen Sahoo.
\newblock Moirai-moe: Empowering time series foundation models with sparse mixture of experts.
\newblock \emph{arXiv preprint arXiv:2410.10469}, 2024{\natexlab{d}}.

\bibitem[Shi et~al.(2024)Shi, Wang, Nie, Li, Ye, Wen, and Jin]{shi2024timemoe}
Xiaoming Shi, Shiyu Wang, Yuqi Nie, Dianqi Li, Zhou Ye, Qingsong Wen, and Ming Jin.
\newblock Time-moe: Billion-scale time series foundation models with mixture of experts, 2024.
\newblock URL \url{https://arxiv.org/abs/2409.16040}.

\bibitem[Dai et~al.(2024)Dai, Deng, Zhao, Xu, Gao, Chen, Li, Zeng, Yu, Wu, et~al.]{dai2024deepseekmoe}
Damai Dai, Chengqi Deng, Chenggang Zhao, Rx~Xu, Huazuo Gao, Deli Chen, Jiashi Li, Wangding Zeng, Xingkai Yu, Y~Wu, et~al.
\newblock Deepseekmoe: Towards ultimate expert specialization in mixture-of-experts language models.
\newblock In \emph{Proceedings of the 62nd Annual Meeting of the Association for Computational Linguistics (Volume 1: Long Papers)}, pages 1280--1297, 2024.

\bibitem[Su et~al.(2024)Su, Ahmed, Lu, Pan, Bo, and Liu]{su2024roformer}
Jianlin Su, Murtadha Ahmed, Yu~Lu, Shengfeng Pan, Wen Bo, and Yunfeng Liu.
\newblock Roformer: Enhanced transformer with rotary position embedding.
\newblock \emph{Neurocomputing}, 568:\penalty0 127063, 2024.

\bibitem[Sun et~al.(2024)Sun, Xie, Eldele, Chen, Hu, and Wu]{sun2024learning}
Yanru Sun, Zongxia Xie, Emadeldeen Eldele, Dongyue Chen, Qinghua Hu, and Min Wu.
\newblock Learning pattern-specific experts for time series forecasting under patch-level distribution shift.
\newblock \emph{arXiv preprint arXiv:2410.09836}, 2024.

\bibitem[Vaswani et~al.(2017)Vaswani, Shazeer, Parmar, Uszkoreit, Jones, Gomez, Kaiser, and Polosukhin]{vaswani2017attention}
Ashish Vaswani, Noam Shazeer, Niki Parmar, Jakob Uszkoreit, Llion Jones, Aidan~N Gomez, {\L}ukasz Kaiser, and Illia Polosukhin.
\newblock Attention is all you need.
\newblock \emph{Advances in neural information processing systems}, 30, 2017.

\bibitem[Wang et~al.(2024)Wang, Gao, Zhao, Sun, and Dai]{wang2024auxiliary}
Lean Wang, Huazuo Gao, Chenggang Zhao, Xu~Sun, and Damai Dai.
\newblock Auxiliary-loss-free load balancing strategy for mixture-of-experts.
\newblock \emph{arXiv preprint arXiv:2408.15664}, 2024.

\bibitem[Hu et~al.(2022)Hu, yelong shen, Wallis, Allen-Zhu, Li, Wang, Wang, and Chen]{hu2022lora}
Edward~J Hu, yelong shen, Phillip Wallis, Zeyuan Allen-Zhu, Yuanzhi Li, Shean Wang, Lu~Wang, and Weizhu Chen.
\newblock Lo{RA}: Low-rank adaptation of large language models.
\newblock In \emph{International Conference on Learning Representations}, 2022.
\newblock URL \url{https://openreview.net/forum?id=nZeVKeeFYf9}.

\bibitem[Jang et~al.(2017)Jang, Gu, and Poole]{jang2017categorical}
Eric Jang, Shixiang Gu, and Ben Poole.
\newblock Categorical reparameterization with gumbel-softmax.
\newblock In \emph{International Conference on Learning Representations}, 2017.
\newblock URL \url{https://openreview.net/forum?id=rkE3y85ee}.

\bibitem[Shazeer et~al.(2017)Shazeer, Mirhoseini, Maziarz, Davis, Le, Hinton, and Dean]{shazeer2017outrageously}
Noam Shazeer, Azalia Mirhoseini, Krzysztof Maziarz, Andy Davis, Quoc~V. Le, Geoffrey~E. Hinton, and Jeff Dean.
\newblock Outrageously large neural networks: The sparsely-gated mixture-of-experts layer.
\newblock In \emph{5th International Conference on Learning Representations, {ICLR} 2017, Toulon, France, April 24-26, 2017, Conference Track Proceedings}. OpenReview.net, 2017.

\bibitem[Vig and Belinkov(2019)]{vig2019analyzing}
Jesse Vig and Yonatan Belinkov.
\newblock Analyzing the structure of attention in a transformer language model.
\newblock \emph{arXiv preprint arXiv:1906.04284}, 2019.

\end{thebibliography}
% \bibliographystyle{unsrtnat}

\end{document}